\documentclass[sigconf,natbib=true]{acmart}

\usepackage[table]{xcolor}
\AtBeginDocument{%
  }

\copyrightyear{2025}
\acmYear{2025}
\setcopyright{cc}
\setcctype{by}
\acmConference[ICBBE 2025]{2025 12th International Conference on Biomedical and Bioinformatics Engineering}{November 27--30, 2025}{Tokyo, Japan}
\acmBooktitle{2025 12th International Conference on Biomedical and Bioinformatics Engineering (ICBBE 2025), November 27--30, 2025, Tokyo, Japan}
\acmPrice{}
\acmDOI{10.1145/3794209.3794211}
\acmISBN{979-8-4007-1965-3/2025/11}

\begin{document}

\title{HistoSeg++: Delving deeper with attention and multiscale feature fusion for biomarker segmentation}


\author{Saad Wazir}
\affiliation{%
  \institution{School of Computing, KAIST}
  \city{Daejeon}
  \country{Republic of Korea}}
\email{saad.wazir@kaist.ac.kr}

\author{Rao Faizan}
\affiliation{%
  \institution{Kyung Hee University (Global Campus)}
  \city{Yongin-si}
  \country{Republic of Korea}}
\email{rao.faizan@khu.ac.kr}

\author{Daeyoung Kim}
\affiliation{%
  \institution{School of Computing, KAIST}
  \city{Daejeon}
  \country{Republic of Korea}}
\email{kimd@kaist.ac.kr}

\renewcommand{\shortauthors}{Wazir et al.}


\begin{abstract}
Segmentation of biomarkers in medical images is frequently viewed as a first step towards medical image analysis in any bioinformatics or biomedical application. Despite progress, existing methods still struggle to capture information at multiple scales and to perform upsampling effectively across different datasets. These shortcomings often result in suboptimal generalization capabilities. Recently, architectures belonging to the Nested-UNet family excel in capturing multiscale contextual information and upsample them effectively. In this work, We propose a novel Nested-UNet architecture that effectively captures multi-scale contextual information. It includes inner and outer attention units to enhance focus during upsampling, along with channel-wise feature recalibration using squeeze-and-excitation modules, leading to improved segmentation performance. Additionally, the architecture integrates an edge-aware loss to emphasize boundary accuracy by assigning greater importance to edge regions. Tested extensively on three publicly available benchmark datasets. Our method demonstrates a generalization performance superior to existing Nested-UNet methods. Code: \href{https://github.com/saadwazir/histosegplusplus}{https://github.com/saadwazir/histosegplusplus}
\end{abstract}



\maketitle

\section{Introduction}
Identifying biomarkers in medical images is a critical first step in many biomedical workflows. It involves examining images to locate important markers. Segmentation of biomarkers within medical imagery is a branch of medical image segmentation that faces difficulties such as scarce sample availability which can hinder semantic segmentation efforts. Notable AI models including UNet \cite{inp:17} and its variants \cite{a:40} improve segmentation performance by employing a nested encoder decoder design with dense skip connections. However these models based on CNN often fail to capture long range dependencies a limitation that has been mitigated by incorporating attention mechanisms \cite{a:6}. Vision Transformers excel at modeling interactions across distant regions yet often struggle to capture precise local details. New architectures such as SegFormer \cite{a:32} and UFormer \cite{inp:20} integrate convolutional modules into transformer frameworks to balance both global and local feature extraction. Even with these advances further refinement is needed to fully model pixel level relationships in every spatial dimension.

To achieve better segmentation accuracy and mitigate the shortcomings, we introduce an architecture that can capture contextual features across multiple scales by leveraging the Nested-UNet arrangements and by leveraging an attention mechanism to effectively upsample features. In this study, our primary focus is to increase the segmentation accuracy of biomarkers, which present numerous challenges due to significant differences in size and appearance and often have unclear boundaries, and to compare our approach with existing Nested-UNet architectures. Below we summarize the main contributions of this study:
\begin{itemize}
    \item We propose a model with a hierarchical arrangement of encoder and decoder modules that efficiently integrates contextual features across multiple scales.
    \item Additionally, we introduced inner and outer attention units, enabling the network to focus on more relevant features during upsampling.
    \item To further enhance the quality of segmentation, we integrated channel-wise feature recalibration using Squeeze-and-Excitation modules.
    \item  We conducted comprehensive experiments on three medical image segmentation datasets and found that our proposed enhancements yield clear performance gains over existing Nested-UNet methods.
\end{itemize}

\begingroup
\renewcommand{\thefootnote}{}
\footnotetext{%
Published in the Proceedings of ICBBE 2025. The Version of Record is available at \href{https://doi.org/10.1145/3794209.3794211}{https://doi.org/10.1145/3794209.3794211}.
}
\addtocounter{footnote}{-1}
\endgroup

\section{Related Work}
 Recent work in \textbf{Convolutional neural network} architectures for semantic segmentation has centered on merging deep features from various levels. Studies \cite{inp:10, inp:11, inp:12, a:39} demonstrate that drawing information from multiple layers enhances overall performance. Variants of UNet which belongs to Nested-UNet family such as U2-Net \cite{a:5} apply multi-scale fusion to capture both local details and broader context. Models like UNet++ \cite{inp:15} and UNet 3+ \cite{inp:19} introduce nested dense and full scale skip connections between encoder and decoder modules and employ deep supervision to strengthen decoder outputs. Despite these improvements, convolutional approaches still find it challenging to model long range dependencies, prompting attention based solutions. For instance MA-UNet \cite{inp:21} integrates multi-scale attention. Even with these advances, convolutional models remain in need of greater flexibility to effectively capture both fine and global information across diverse data sets. \textbf{Transformer-based architectures} represent a significant advancement in computer vision. Initially designed to implement complex attention in natural language processing they have also been adapted for image based tasks. In segmentation TransUNet \cite{a:21} employs a transformer encoder to learn rich feature representations along with a UNet style CNN-based decoder to upsample those features effectively. Swin-UNet \cite{inp:9} utilises a hierarchical Swin Transformer encoder with shifted windows to gather multi-scale context and a matching decoder with a patch expanding layer to restore spatial detail during upsampling. However most of transformer-based methods depend on large data sets to capture long range dependencies.

\begin{table}[]
\centering
\caption{Quantitative results of the proposed work and comparison of existing methods. The best results are highlighted in bold, while the second best results are underlined.}
\label{tab:tab-all-sp}
\resizebox{\columnwidth}{!}{%
\begin{tabular}{llllll}
\hline
\textbf{Model} &
  \textbf{IoU $\uparrow$} &
  \textbf{Dice $\uparrow$} &
  \textbf{Prec. $\uparrow$} &
  \textbf{Rec. $\uparrow$} &
  \textbf{HD95 $\downarrow$} \\
\hline
\rowcolor[HTML]{EFEFEF} \multicolumn{6}{c}{\textbf{MoNuSeg}} \\ \hline
UNet       & 59.62  & 73.42  & 74.87  & 73.04  & 3.6609      \\
HoVer-Net  & 62.90  & 77.26  & 77.67  & 77.54  & 3.8687      \\
nnU-Net    & 67.60  & 80.42  & \textbf{81.63}  & 80.88  & 3.3357 \\
UNet++     & \underline{69.34}  & \underline{81.83}  & 75.31  & \underline{90.29}  & \underline{2.7901}      \\
UNet 3+    & 69.05  & 81.58  & 77.79  & 86.71  & 2.8740      \\
U2-Net     & 68.89  & 81.33  & 77.20  & 86.88  & 13.445      \\
Swin-UNet  & 65.38  & 79.01  & 72.38  & 87.88  & 3.7871      \\
TransUNet  & 67.89  & 80.22  & \underline{81.58}  & 79.17  & 3.5602      \\
Ours       & \textbf{71.44}  & \textbf{83.30}  & 76.12  & \textbf{92.33}  & \textbf{2.4399} \\ \hline

\rowcolor[HTML]{EFEFEF} \multicolumn{6}{c}{\textbf{DSB}} \\ \hline
UNet       & 83.83  & 90.17  & 89.09  & \underline{93.78}  & \underline{7.6797}      \\
HoVer-Net  & 81.29  & 89.04  & 88.73  & 91.01  & \textbf{7.6289}      \\
nnU-Net    & 77.24  & 86.09  & 90.18  & 84.33  & 9.9094 \\
UNet++     & 83.95  & 91.10  & 88.49  & \textbf{94.26}  & 10.045      \\
UNet 3+    & 69.11  & 75.76  & 78.48  & 77.49  & 19.272      \\
U2-Net     & 82.96  & 90.03  & 90.28  & 91.08  & 10.869      \\
Swin-UNet  & 84.49  & \underline{91.03}  & 90.77  & 92.72  & 8.0415 \\
TransUNet  & \underline{84.72}  & 90.90  & \underline{93.37}  & 89.86  & 11.213      \\
Ours       & \textbf{85.50}  & \textbf{91.81}  & \textbf{95.84}  & 88.90  & 8.8634      \\ \hline

\rowcolor[HTML]{EFEFEF} \multicolumn{6}{c}{\textbf{Electron Microscopy}} \\ \hline
UNet       & 77.45  & 85.37  & 88.82  & 84.95  & 46.803      \\
HoVer-Net  & 79.18  & 88.31  & 91.43  & 85.55  & 12.624      \\
nnU-Net    & 85.02  & 91.86  & \textbf{94.75}  & 89.23  & \underline{5.2735}      \\
UNet++     & \textbf{87.01}  & \underline{92.49}  & 92.53  & \underline{93.38}  & \textbf{4.7633} \\
UNet 3+    & 80.35  & 89.03  & 92.53  & 86.02  & 8.8756      \\
U2-Net     & 86.31  & 92.01  & 91.92  & 92.88  & 20.817      \\
Swin-UNet  & 83.72  & 91.08  & 92.18  & 90.10  & 9.7707      \\
TransUNet  & 84.86  & 91.77  & \underline{94.33}  & 89.42  & 6.1746 \\
Ours       & \underline{86.54}  & \textbf{92.74}  & 91.16  & \textbf{94.50}  & 6.4787      \\ \hline

\end{tabular}%
}
\end{table}

\section{Method}

\begin{figure*}
\centerline{\includegraphics[width=1\textwidth]{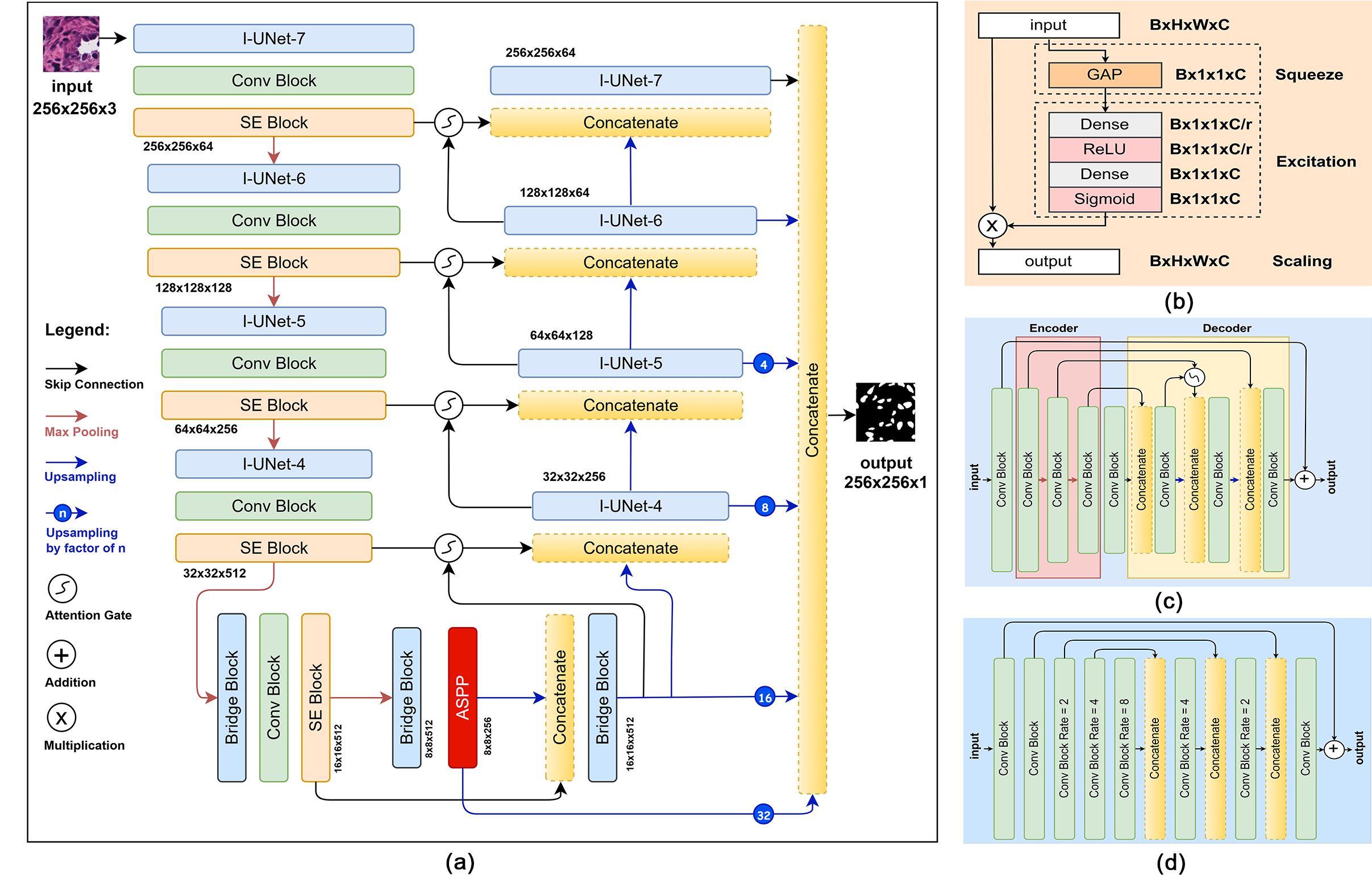}}
\caption{An overview of the proposed architecture (a) Outer-UNet (b) SE Block. (c) I-UNet Block with k = 4 (d) Bridge Block.}
\label{fig:arch}
\end{figure*}

\subsection{Overview}
This architecture employs a hierarchical nested design in which each Inner-UNet module refines feature representations at a particular resolution by combining convolution blocks with attention gates. These inner modules are embedded within a larger Outer-UNet scaffold that manages the progression of information through successive downsampling and upsampling stages. The outer network draws on the outputs of all inner modules to assemble multi-scale feature maps, ensuring that both local details and global context are maintained throughout the segmentation process. By integrating channel recalibration at every stage, the model delivers strong feature encoding and precise mask generation across diverse medical imaging data sets.

\subsection{Squeeze and Excitation (SE) Block:} As described in \cite{inp:1} and illustrated in Figure \ref{fig:arch} part b this block first compresses feature maps using global average pooling then applies dense layers with ReLU and sigmoid activations to compute channel-wise scaling factors. These factors are multiplied with the original feature maps to enable channel-wise attention.

\subsection{Convolution Block:} This block applies depthwise separable convolution layers followed by batch normalization and a LeakyReLU activation to capture and refine feature representations and to enable non linear transformations.

\subsection{Attention Gate:} To prioritize key features from skip connections, we use an attention mechanism following \cite{a:6}. The attention mechanism filters out irrelevant and noisy signals before combining important activation. This mechanism also regulates gradient flow to focus updates on crucial spatial areas, enhancing model accuracy.

\subsection{Atrous Spatial Pyramid Pooling (ASPP):} As introduced in Deeplab \cite{a:7}, the ASPP block maintains the spatial dimension of feature maps while extracting high-semantic features with a larger field-of-view using the multi-grid method. We have modified it to use depthwise separable convolution.

\subsection{Inner-UNet (I-UNet) Block}
The I-UNet block shown in Figure \ref{fig:arch} part c is designed to capture long range and multiple scale contextual information through an attention gate. Convolution blocks followed by max pooling in the encoder extract hierarchical features, and a bridge convolution block connects this representation to the decoder. During decoding upsampling layers work together with the attention gate and convolution blocks to merge coarse and fine details via skip connections. A final residual connection then fuses the original local feature maps with the aggregated long range multiple scale information to produce the block’s output. Various configurations of the I-UNet module are embedded within the Outer-UNet encoder and decoder and are designated by a parameter k. The value of k combines the initial convolution block with the total number of convolution blocks in the encoder. In this setup k minus one corresponds to the count of encoder convolution blocks, k minus two indicates both the number of max-pooling layers in the encoder and upsampling layers in the decoder, and twice k gives the total convolution block count inside the I-UNet module. All variants share a symmetric design that incorporates attention gates and skip connections. Figure \ref{fig:arch} part c depicts the I-UNet 4 variant.

\subsection{Bridge Block}
In the deeper layers, if the spatial dimension becomes significantly small, the network cannot extract semantic information. Therefore, we use dilated convolutions to allow the network to learn dense semantic information. As shown in Figure \ref{fig:arch} part d.
\subsection{Outer-UNet}
The core network appears in Figure \ref{fig:arch} part a and brings together all the components described above. Its aim is to extract feature maps, merge them effectively and then recover spatial resolution. The encoder path is built from four Inner-UNet modules named I-UNet 7, 6, 5, and 4 each followed by a convolution block for spatial feature learning an SE block for channel-wise recalibration and a max pooling layer for downsampling. The bridge block then refines these deep representations and the ASPP block gathers multi-scale context. In the decoder path upsampling layers, attention gates and skip connections reverse the process to fuse encoder features precisely. After each upsampling step one of the Inner-UNet modules is applied. Finally outputs from all decoder stages are resized to the original resolution concatenated and passed through a 1x1 convolution before a sigmoid activation produces the final binary mask. This design enables the model to learn a variety of spatial features and blend them seamlessly to create segmentation masks that cover multiple scales and contexts.

\section{Experiments}

\subsection{Datasets}

To evaluate our approach we used three publicly available segmentation datasets. These sets include images captured under varying illumination conditions and magnifications and feature different cell types and imaging modalities. Our goal was to assess the model’s ability to generalize across these diverse scenarios. For training we generated overlapping patches of size 256 × 256 pixels with a stride of 128. During testing, we applied the same patch generation procedure to perform predictions.

\textbf{Data Augmentations:} We apply both offline and online augmentation during training. Using offline augmentation on the training sets we generated 8820 samples for MoNuSeg, 13710 samples for DSB, and 9852 samples for the EM dataset. Offline transforms are applied with full certainty and consist of random brightness and contrast adjustments grid distortion image transposition and elastic transformation. During online data loading we perform at each iteration rotations up to 90 degrees width and height shifts of up to 0.3 of the image dimension shear operations of up to 0.5 zoom of up to 0.3 and random horizontal and vertical flips.\\\\
The details of the datasets are as follows:
\begin{enumerate}
\item \textbf{The Multi-organ Nucleus Segmentation (MoNuSeg)} \cite{a:1} dataset contains H\&E-stained histopathology images. The dataset comprises 44 images, each with dimensions of 1000 by 1000 pixels. The MoNuSeg dataset consists of 30 images allocated for training and an additional 14 images reserved for testing.\\
\item \textbf{The 2018 Data Science Bowl (DSB)} \cite{a:3} dataset encompasses a diverse collection of segmented cell images acquired under various conditions and from different organisms. The images have dimensions ranging from 128 x 128 to 512 × 512 pixels. We solely relied on the stage 1 training set due to the unavailability of publicly accessible ground truth masks.\\
\item \textbf{Electron Microscopy (EM) } \cite{inp:24} dataset features annotated mitochondria from the CA1 hippocampus region of the brain. The training and testing sets each contain 165 images with corresponding masks, with a size of 768x1024 pixels.
\end{enumerate}

\subsection{Evaluation Metrics}

We evaluated both existing deep learning methods and our HistoSeg++ using standard evaluation metrics  \cite{a:12}. Pixel based performance was assessed using IoU, Dice, precision and recall. To capture surface distance characteristics we included the Hausdorff distance at the ninety fifth percentile which enriches our evaluation. Each metric was calculated for every prediction and then averaged across all test samples.

\subsection{Experimental Setup}
We trained the network with the Adam optimizer using a learning rate of 0.0001 and a batch size of eight and initialized convolution kernels with the He Uniform scheme. Training was carried out from scratch for two hundred epochs. For MoNuSeg and EM we used their standard test sets while for DSB we adopted a nine to ten split between training and evaluation data. We created a validation set from the training samples, selected the checkpoints that performed best on the validation set, and used them for evaluation. We reproduce all the results for existing methods using their publicly available repositories and use the optimal settings reported in their work. For our work, we employed the edge-aware loss function which improves boundary accuracy in segmentation by first computing cross-entropy between predicted and ground truth masks and then amplifying errors at true object edges. Edges are identified via Sobel filtering followed by a gradient magnitude threshold. Error values at those edge locations are multiplied by a factor greater than one before averaging across all pixels.\textbf{ Edge-aware loss function} is defined as:
\begin{equation}
L(y, \hat{y}) = \frac{1}{N} \sum_{i=1}^N \left(1 + \left(\text{em}_i \times (w - 1)\right)\right) \times \text{BCE}(y_i, \hat{y}_i)
\end{equation}
where
\[
\text{edge\_mask (em)} = \frac{1}{D} \sum_{d=1}^D \mathbf{1}\left(\sqrt{s_x^2 + s_y^2} > 0.1\right),
\]
\[
s_x = \text{Sobel}_x(y), \quad s_y = \text{Sobel}_y(y),
\]
and \( w \) is the edge weight, \( N \) is the number of samples, and \( D \) is the spatial dimensionality of the mask.

\subsection{Results}
\begin{figure}
\centerline{\includegraphics[width=0.5\textwidth]{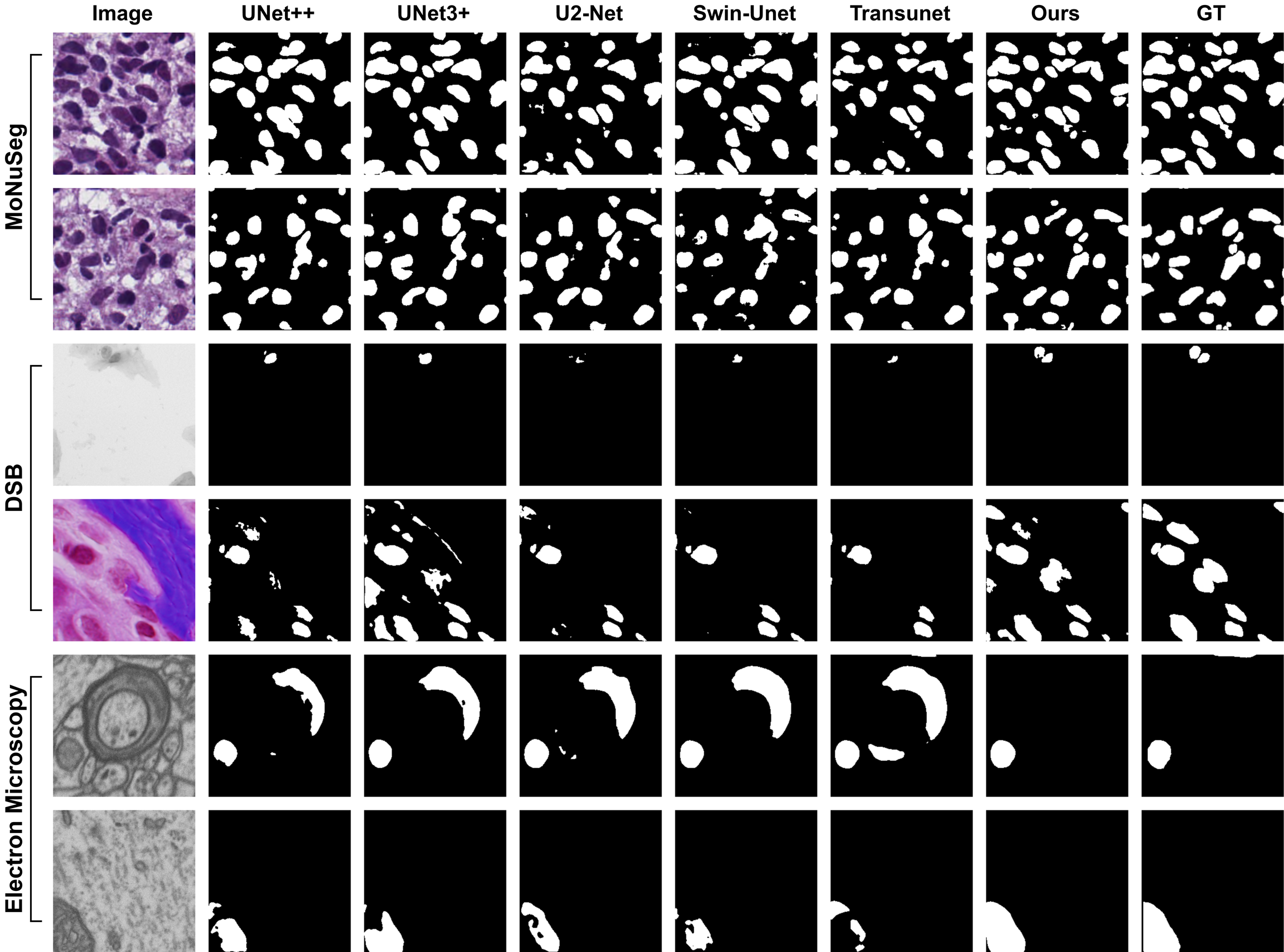}}
\caption{Qualitative Results Comparison}
\label{fig:qual}
\end{figure}

To evaluate our proposed method and existing methods, we selected baseline models such as UNet \cite{inp:17} and nnU-Net \cite{a:34}, as they are original architectures widely used in segmentation tasks. HoVer-Net \cite{a:36} is included because it serves as a baseline for biomarker segmentation. Swin-UNet \cite{inp:9} and TransUNet \cite{a:21} are baseline transformer-based methods for segmentation. UNet++ \cite{inp:15}, UNet 3+ \cite{inp:19}, and U2-Net \cite{a:5} are SOTA methods from the Nested-UNet family.
\textbf{Quantitative comparison} is presented in Table \ref{tab:tab-all-sp}. For the nuclei segmentation, HistoSeg++ achieves the highest accuracy, outperforming transformer-based methods such as Swin-UNet, which exhibit lower segmentation fidelity. Among convolutional variants, UNet++ and nnU-Net deliver solid results but lag in capturing fine nuclear details. In the cell segmentation dataset, characterized by diverse morphologies and imaging conditions, our approach again leads in overall quality and precision. Transformer backbones like TransUNet show strong precision but often miss subtle cell regions, while UNet 3+ performs weakest, struggling with overlap accuracy and boundary localization. For mitochondrial segmentation, the nested skip connections and deep supervision of UNet++ enable strong multi-scale feature extraction. In contrast, the original UNet performs poorly, reflecting its limited capacity for modeling long-range dependencies in complex micrographs. By combining edge-aware loss and nested Inner-UNet modules, our method yields the most complete and precise mitochondrial masks.
\textbf{Qualitative results} are shown in Figure \ref{fig:qual}. Transformer-based methods often over-segment and add artifacts, while Nested-UNet variants yield precise, coherent masks. Our approach produces reliable segmentation with minimal over-segmentation.

\subsection{Ablation Studies}

\begin{table}{}
\centering
\caption{Design choices and their impacts in the proposed approach.\\}
\label{tab:abb-modules}
\resizebox{0.45\textwidth}{!}{%
\begin{tabular}{lcc}
\hline
\textbf{Model} & \textbf{IoU $\uparrow$} & \textbf{HD95 $\downarrow$} \\ \hline
Baseline                                          & 66.05 & 3.1266 \\
Baseline + in\_A                                  & 68.28 & 3.0010 \\
Baseline + out\_A                                 & 68.95 & 2.7103 \\
Baseline + in\_A + out\_A                         & 69.15 & 2.5767 \\
Baseline + in\_A + out\_A + ASPP                  & 71.12 & 2.5557 \\
Baseline + in\_A + out\_A + ASPP + SE             & 71.44 & 2.4399 \\ \hline
\end{tabular}%
}
\end{table}

We conducted ablation studies to assess the effectiveness of components in our model, using the MoNuSeg dataset for the experiments, as detailed in Table \ref{tab:abb-modules}. Our baseline involves nested U-Nets in an encoder-decoder setup. Adding inner (in\_A) and outer (out\_A) attention improves all metrics, notably enhancing Recall, which aids in better identifying relevant cases. Integrating ASPP, and SE modules boosts IoU and Dice scores for increased segmentation accuracy. These advancements consistently improve segmentation but require more computational resources.

\section{Conclusion}
This work presents a novel deep neural framework for precise segmentation in medical imaging. The proposed model combines receptive fields of varying sizes with spatial and channel attention modules to capture contextual cues at multiple scales and improve feature reconstruction during upsampling. We validate its performance on three public datasets, demonstrating consistent gains in segmentation quality. With further refinement the approach could enable real time processing and support multi class segmentation.

\section*{Acknowledgments}
This work was supported by the National Research Foundation of Korea(NRF) grant funded by the Korea government(MSIT)(RS-2025-00573160).


\bibliographystyle{ACM-Reference-Format}
\bibliography{software}

\end{document}